# Fast low-level pattern matching algorithm

Janja Paliska Soldo[a], Ana Sović Kržić[b*] and Damir Seršić[c]

[a]*HashCode, Frana Folnegovica 1c, 10000 Zagreb, Croatia, janjapaliska@gmail.com*

[b]*University of Zagreb Faculty of Electrical Engineering and Computing, 10000 Zagreb, Croatia, ana.sovic.krzic@fer.hr*

[c]*University of Zagreb Faculty of Electrical Engineering and Computing, 10000 Zagreb, Croatia, damir.sersic@fer.hr*

∗*Corresponding author. Tel.: +385-1-6129-883; fax: +385-1-6129-652*

*ABSTRACT*

This paper focuses on pattern matching in the DNA sequence. It was inspired by a previously reported method that proposes encoding both pattern and sequence using prime numbers. Although fast, the method is limited to rather small pattern lengths, due to computing precision problem. Our approach successfully deals with large patterns, due to our implementation that uses modular arithmetic. In order to get the results very fast, the code was adapted for multithreading and parallel implementations. The method is reduced to assembly language level instructions, thus the final result shows significant time and memory savings compared to the reference algorithm.

# 1. Introduction

Pattern matching problem is a frequent problem in various fields of science. In bioinformatics, it has large usage for the DNA sequencing, which is an important task in genomics.

*1.1. Pattern matching problem*

Exact pattern matching problem, in general, finds all subsets of text $t$ that are equal to the given pattern $p$. Let us denote text length as $n$ and pattern length as $m$ ($n > m$). Both $t$ and $p$ are strings over a finite alphabet $\Sigma = \{a_1, \ldots, a_\sigma\}$. Pattern $p$ is said to be found on location $i$ in $t$ if $\forall 1 \leq j \leq m$, $p[j] = t[i+j-1]$ is fulfilled [1].

Pattern matching with character classes allows $p$ to be consisted of character classes such that $p[i] \subseteq \Sigma$. Here, $p$ is found on locations $i$ if $\forall 1 \leq j \leq m$, $t[i+j-1] \in p[j]$ is fulfilled. For example, pattern $p = b[abc]cc[bc]$ in given text $t =$ '*accabbccba*' occurs on location 5 (substring '*bbccb*').

Approximate pattern matching extends the problem so that there are at most $k$ allowed mismatches between $t$ and $p$. In the previous example, if we set $k = 2$, there are more occurrences of $p$ in $t$ (mismatches are underlined): location $i=1$ (substring '*accab*'), location $i=4$ (substring '*abbcc*') and location $i=6$ (substring '*bccba*'). To find all occurrences of the pattern in a given text, it is usually scanned by the sliding window of pattern length size $m$ [2].

*1.2. Literature overview*

The brute force algorithm for exact pattern matching has computational time $O(mn)$. In the worst case, $m$ characters at each of the $n$ letters of the text must be checked.

Our paper is inspired by work of Linhart and Shamir [1]. More details are given in paragraph 1.3. Here, we present a brief overview of other competing pattern matching methods.

Boyer-Moore algorithm is an exact, single pattern matching method. It starts with the last character of the pattern and aligns it to the first appearing in text. If the mismatch between other characters is detected, pattern is shifted right so the detected mismatch character is aligned to the right most occurrence of it in the pattern. The algorithm uses bad character rule and good suffix rule. It has the worst case complexity $O(m+n)$ and of the best case $O(n/m)$ [3]. Horspool simplified the Boyer-Moore algorithm. It uses only the bad character rule on the rightmost character of the window to compute the shifts in the Boyer-Moore algorithm [4]. The Sunday's quick-search algorithm generates bad character shift table during the preprocessing stage. During the search algorithm, the pattern symbol can be compared in an arbitrary order [5].

Sheik's algorithm reduces character comparison. The last character of the window and the pattern is compared. If there is a match, the algorithm compares the first character of the window to the pattern. The remaining characters are compared from right to the left until mismatch or complete match occurs. The window shift is provided by the quick-search bad character placed next to the window [6]. Time complexity in the best case is $O(n/(m+1))$, and in the worst case is $O(m(n-m+1))$. The preprocessing phase time complexity is $O(m+\sigma)$ and the space complexity is $O(\sigma)$, where $\sigma$ is the alphabet size.

Bhukya and Somayajulu create an array of indices of character pairs to reduce the number of comparisons, as well as an array of frequencies of each pair. Index table for the pattern is created in the same way. Search starts with a pair in the text with the least frequency. By using the index, the possible location of the pattern in the sequence is found. The rest of the pattern is compared sequentially [7].

A bit different approach is using keyword and suffix trees. Keyword tree is built of patterns. Each edge of the keyword tree is labeled with a letter from an alphabet. Every stored keyword can be spelled on a path from root to some leaf and every path from root to leaf gives a keyword. Construction of the keyword tree has computational time proportional to the combined lengths of the patterns. The text is threaded through the keyword tree [8]. Suffix tree is similar to keyword tree, except each edge is label with a substring of the text. All internal edges have at least two outgoing edges. The method has computational time $O(n^2+n)$ [8]. Generalized faster suffix trees methods that allow the detection with $k$ mismatches have computational time $O(kn)$ [8].

Approximate pattern matching methods can be divided onto methods based on dynamic programming matrix, automata, filtering and bit-parallelism [9].

One of the first dynamic programming methods for pattern matching uses an evolutionary distance metric. Computational time to recognize all pairs of intervals is $O(mn)$, since any text position is the potential start of a match [10]. Other distance measures that are used are Hamming distance and Levenshtein distance. Landau and Vishkin compute the dynamic programming matrix diagonal-wise instead of column-wise and use the Hamming distance. The computational time is $O(k^2n)$ [11]. The Levenshtein distance is the minimum number of character changes, insertions and deletions required to transform one string into other. The strings are not necessarily of equal length [12]. Many authors tried to decrease the worst case computational time, e.g. Galil and Park [13]. Their algorithm consists of preprocessing of the pattern (building the matching statistics of the pattern against itself) and processing of the text. It has computational time $O(km)$.

Ukkonen proposed deterministic automaton for pattern matching, where each possible set of values for the columns of the dynamic programming matrix is a state of the automaton [12]. A huge number of states are reduced by replacing all the elements in the columns that are larger than $k+1$ by $k+1$ without affecting the output of the search.

Filtering methods invert the problem: they tell if a text position does not match a pattern. Although the filtering methods are the fastest, they need a non-filter algorithm to check the potential matches. Most filtering approaches search pieces of the pattern without errors, since the exact algorithms are faster [9].

Bit-parallelism algorithms are very efficient for any error level and especially when short patterns are involved. Usually, parallelism is used with non-deterministic automaton [14] and dynamic programming matrix. The algorithms compute bit operations: or, and, xor, not and shifting, as well as arithmetic operations on the bits: adding and subtracting [9][15].

Although, there are numerous existing solutions for pattern matching problem, not all of them are appropriate for usage in bioinformatics. In bioinformatics, the alphabet consists of four letters, $\Sigma = \{A, C, G, T\}$, corresponding to the four nucleobases: adenine (A), cytosine (C), guanine (G) and thymine (T). The most popular pattern matching methods are heuristic similarity search methods: MAFFT, FASTA and BLAST. MAFFT detects homologous regions using the fast Fourier transform. The method has high accuracy of alignments even for sequences with large insertions or extensions [16][17]. FASTA identifies regions of similar sequences and scores the aligned identical and differing residues in those regions. It uses the Smith-Waterman algorithm for calculating an optimal score for the alignment [18][19]. BLAST combines heuristic approach and the Smith-Waterman algorithm. The main differences between BLAST and FASTA are in making the list of the possible matching words. FASTA cares about all of the common words in the pattern and the text and BLAST only cares about the high-scoring words [20]. A GraphMap method aligns long reads with speed and high precision using five stages: region selection, graph-based vertex-centric construction of anchors, extending anchors into alignments, refining alignments using L1 linear regression and construction of final alignment [21].

Although heuristic methods are very fast, they cannot guarantee the optimal pattern search. Our proposed method finds all occurrences of the pattern in a given text allowing predefined *k* mismatches. Paper that motivated our algorithm uses prime number encoding and Fast Fourier Transform (FFT) [1]. It successfully deals with character classes instead of pure characters, as well with exact and non-exact matches. Main problem is dealing with large biological patterns. Our proposed algorithm provides a fast implementation of the same principle that is far more appropriate for large data sets.

*1.3. Pattern matching with character classes using prime number encoding*

In [1], Linhart and Shamir presented a method that is based on prime number encoding and computing convolution between the encoded pattern and text using FFT. Pattern and text symbols are encoded differently. Here, we give a brief explanation.

Each symbol $a_i \in \Sigma$ is assigned to a different prime number $k_i$, where all prime numbers are larger than pattern length *m*: $m < k_1 < k_2 < \ldots < k_\sigma$.

*1.3.1. Text encoding*

Let *M* be a product of σ prime numbers, $M = k_1 \cdot k_2 \cdot \ldots \cdot k_\sigma$. Text contains only one-letter symbols. Each symbol $a_i \in \Sigma$ in the text is encoded by the integer $e_i = M / k_i$.

*1.3.2. Pattern encoding*

Each character class $[a_{i1}, \ldots, a_{ic}]$ in the pattern is encoded by integer $n_S$, where $S = \{i_1, \ldots, i_c\}$, and it is computed with the system of linear congruencies:

$$n_S \equiv \begin{cases} 0 \ (\mathrm{mod}\, k_i), \forall i \in S, \\ 1 \ (\mathrm{mod}\, k_j), \forall j \notin S. \end{cases} \quad (1)$$

Chinese reminder theorem (CRT) guarantees that it is always possible to compute exactly one integer in the range [0, $M$⟩ so the system (1) is satisfied.

### 1.3.3. Convolution

After each symbol of the text and each symbol of the pattern have been encoded, the convolution between them is calculated:

$$(p * t)[i] = \sum_{j=-\infty}^{\infty} p[j] \cdot t[i-j], \forall i \in \mathbb{Z}. \quad (2)$$

The convolution is usually calculated using the FFT:

$p_{encoded} * t_{encoded} = \mathrm{IFFT}\ (\mathrm{FFT}(p_{encoded}) \cdot \mathrm{FFT}\ (t_{encoded}))$.

### 1.3.4. Matching

Finally, exact matches are found on every position $i$ where

$$(p_{encoded} * t_{encoded})[i] \bmod M = 0. \quad (3)$$

### 1.3.5. Problems and limitations

The main problem with this algorithm is that the smallest prime number has to be greater than the pattern length. If it is used in the DNA sequencing, data sequences are large, and the algorithm has limitations with computational precision.

For example, we need four prime numbers for coding symbols from alphabet $\Sigma$ = {A, C, G, T}. If the pattern length is $m$=1000, the smallest prime numbers are: $k_1$=1009, $k_2$=1013, $k_3$=1019 and $k_4$=1021. The product is $M = k_1 \cdot k_2 \cdot k_3 \cdot k_4 \approx 10^{12}$. Each symbol $a_i$ in the text is encoded by the integer $M / k_i \approx 10^9$. Each character class [$a_{i1}$,…, $a_{ic}$] in the pattern is encoded by integer $n_S$, that can be any number in range [0 – $10^{12}$]. Next step is convolution, where each text symbol (order ≈ $10^9$) is multiplied with each pattern symbol (order ≈ $10^{12}$). There is a danger to exceed the greatest provided data range (e.g. uint64 ≈ $1.8 \cdot 10^{19}$).

Given example was for pattern length $m$ = 1000. But, in modern DNA sequencing, lengths often exceed ten thousand samples. Therefore, it is necessary to adjust this method for use with large numbers. In the next chapter we propose an efficient, quick and exact algorithm.

## 2. Proposed Pattern Matching in Bioinformatics

The final matching depends only on remainder after division with *M*, as given in (3). This leads to the fact that it is not necessary to carry out the multiplication of large numbers, but we can observe the remainders or the matching pairs of the symbols, both in text and pattern.

*2.1. Text encoding*

A basic alphabet in the DNA is Σ = {A, C, G, T}. Since we have only four possible symbols, each text symbol is encoded by two-bit codes shown in Table 1.

*2.2. Pattern encoding*

Using four symbols and according to IUB/IUPAC standard [22], we have 15 possible character classes denoted with a single-letter code. Each pattern symbol is encoded with four-bit binary codes (Table 2). If the symbol is missing or has error, it is coded with $1111_{(2)}$.

*2.3. Modular arithmetic*

Modular arithmetic is implemented in such a way that for every possible pair of the text and the pattern symbol it assigns value 0 for a match and 1 for a mismatch. For example, pair A – M ([AC]) has value 0 since it is a match, while pair A – B ([CGT]) has value 1 since it is a mismatch. All possible values are provided in Table 3.

Numbers in brackets represent an order number of each pair. Table 3 is implemented as a dictionary. Values 1 and 0 are written in a one dimensional integer array of size 64, and unique key to approach certain value (array index) is obtained from pattern and text encoded symbols as it will be explained in the sequel. Value in the brackets matches the array index where certain 1 or 0 is located.

After encoding text and the pattern, the text is compared with the pattern. For each text symbol $t[i]$ and pattern symbol $p[j]$, we calculate index in the integer array:

1. shift the pattern symbol code to the left for two locations: $r[j] = p[j] << 2$, where symbol << represents shifting to the left of the bits,

2. calculate OR operation between shifted pattern symbol code and text symbol code: $l_{(2)} = r[j]$ OR $t[i]$. Given $l_{(10)}$ is the array index from the Table 3.

3. by reading from the Table 3 at the index $l_{(10)}$, we find if there is match or mismatch between the text and pattern symbol.

In other words, with two low level operations (in assembler: shift left by 2 followed by OR) and reading from the dictionary, we obtain all pair values very quickly.

For example, if we compare $t[i] = A$ and $p[j] = D$, we have text code A = 00, and pattern code D = 1100. Index of the pair matching is calculated using shift and OR operation (1100 << 2) OR (00) = 110000 OR 00 = $110000_{(2)} = 48_{(10)}$. Value at the position 48 in the Table 3 is 0, which corresponds to match. If we look at the Table 3 and search for the value assigned to pair A – D, we will find 0.

*2.4. Matching*

We define *k* as a number of the allowed mismatches between *t* and *p*. Match at the location *i* is found if:

$$\sum_{j=1}^{m} \mathbf{L}\big[(p[j] \ll 2) \text{OR } t[i+j-1]\big] \leq k, \qquad (4)$$

where **L** is the dictionary from the Table 3. If no mismatches are allowed, the above summation stops and reset when the first 1 comes across.

*2.5. Example*

Let us consider a search for the pattern *p* = C [CGT]GG[CG] in the text *t* = ATGACCGGCAT. We allow *k* = 2 mismatches.

**STEP 1.** We need to code the text using Table 1:

$$t_{(10)} = [0 \quad 3 \quad 2 \quad 0 \quad 1 \quad 1 \quad 2 \quad 2 \quad 1 \quad 0 \quad 3],$$

or in binary numeral system:

$$t_{(2)} = [00 \quad 11 \quad 10 \quad 00 \quad 01 \quad 01 \quad 10 \quad 10 \quad 01 \quad 00 \quad 11].$$

**STEP 2.** Code the pattern using Table 2:

$$p_{(10)} = [1 \quad 13 \quad 2 \quad 2 \quad 7],$$

or in binary numeral system:

$$p_{(2)} = [0001 \quad 1101 \quad 0010 \quad 0010 \quad 0111].$$

**STEP 3.** Each element of vector $p_{(2)}$ is shifted to the left by two locations:

$$r_{(2)} = [000100 \quad 110100 \quad 001000 \quad 001000 \quad 011100].$$

**STEP 4.** Calculate matching using modified convolution between text and all possible shifts of the pattern (Table 4). Firstly, without shift, e.g. *I* = 1 (first part of the Table 4).

OR operation between each element of the $r_{(2)}$ and first 5 elements of the $t_{(2)}$ is calculated: $t_{(2)}[1]$ OR $r_{(2)}[1] = 00_{(2)}$ OR $000100_{(2)}$ = $000100_{(2)}=4_{(10)}$. On 4[th] position in the dictionary **L** (Table 3) is 1. That means a mismatch. If we allow zero mismatches, checking for position *i* = 1 of the pattern would stop here.

Since we allow 2 mismatches, we observe second elements of the text and of the pattern: $t_{(2)}[2]$ OR $r_{(2)}[2] = 11_{(2)}$ OR $110100_{(2)} = 110111_{(2)} = 55_{(10)}$. Based on the dictionary **L**, a match is found.

Furthermore, we observe third elements of the text and of the pattern: $t_{(2)}[3]$ OR $r_{(2)}[3] = 10_{(2)}$ OR $001000_{(2)} = 001010_{(2)} = 10_{(10)}$. Based on the dictionary **L**, a match is found.

Now we observe fourth elements of the text and of the pattern: $t_{(2)}[4]$ OR $r_{(2)}[4] = 00_{(2)}$ OR $001000_{(2)} = 001000_{(2)} = 8_{(10)}$. Based on the dictionary **L**, a match is not found. Cumulatively, we have two mismatches. If we allow only one mismatch, checking for this position of the pattern would stop here. However, we allowed two mismatches; therefore, we need to check the last element of the pattern.

OR operation between the fifth elements of the text and of the pattern: $t_{(2)}[5]$ OR $r_{(2)}[5] = 01_{(2)}$ OR $011100_{(2)} = 011101_{(2)} = 29_{(10)}$. Based on the dictionary **L**, a match is not found. Cumulatively, we still have two mismatches.

We reached the end of the pattern and only two mismatches were found. We conclude that we found our pattern in the given text with the allowed number of the mismatches.

We shift the pattern to the right, and compare the first element of the pattern with the second element of the text (second part of the Table 4). We found a mismatch. Comparing the second element of the pattern and the third element of the text we found a match. The third element of the pattern and the fourth element of the text give a mismatch. Finally, the fourth element of the pattern and the fifth element of the text give a mismatch. Since we have 3 mismatches, calculations of the OR operations and using dictionary are over. For position $i = 2$, we do not have a match in the allowed limits.

We slide the pattern successively to the right and repeat the same algorithm. After all possible shifts, we get: perfect match on the position $i = 5$ and matches with 2 allowed mismatches at the positions: $i = 1$, $i = 4$ and $i = 6$.

*2.6. Further acceleration by implementing parallelization*

This algorithm is very suitable for parallelization with multithreading (running on a cluster of computers). It is simply necessary to divide encoded text to $n/numberOfThreads$ pieces with $m$-1 overlaps between consecutive pieces and search every piece with its own thread.

**3. Experiments**

Proposed method is tested on the human genome, chromosome sequence 2, size 243199373 and name Homo_sapiens.GRCh37.72.dna.chromosome.2.fa [23]. Pattern size $m$ are 500, 10 000 and 100 000 on 5 different position in the referent DNA. The results are presented in Table 5, Table 6 and Table 7. Each pattern is tested for 5 times and mean value of the runtimes is calculated.

We see, from the measured runtimes, that the sizes of the pattern do not extended the runtime. For matching a pattern of size $m$=500, the algorithm needs 6.595 seconds; for the pattern size $m$=10000, it needs 6.882 seconds; and for the pattern size

$m$=100000, it needs 6.980 seconds. The largest pattern is 200 times bigger than the smallest one and the runtime is only 1.0584 times longer.

## 4. Conclusion

In this paper, we present an improvement of Linhart and Shamir pattern matching method with character classes using prime number encoding [1]. For long patterns, their approach demands arithmetic of a very high precision, which sets practical limits on the pattern length for the method. In our approach, characters from text are encoded with two-bit codes and character classes from pattern are encoded with four-bit codes. Using modular arithmetic and reading the results from dictionary, a perfect match or matches with allowed number of mismatches are found. Since low-level bit operations are used, the method is very fast. Further acceleration is done by parallelization. Our approach does not put any limits on the pattern length thus over performing the competitive methods.

**Acknowledgments**

This work has been fully supported by Croatian Science Foundation under the project UIP-11-2013-7353 "Algorithms for Genome Sequence Analysis".

**Table 1. Text symbol codes.**

| Symbol | Code (binary) | Value (decimal) |
|--------|---------------|-----------------|
| A      | $00_{(2)}$    | $0_{(10)}$      |
| C      | $01_{(2)}$    | $1_{(10)}$      |
| G      | $10_{(2)}$    | $2_{(10)}$      |
| T      | $11_{(2)}$    | $3_{(10)}$      |

**Table 2. Character classes and pattern symbol codes.**

| Character class | Symbol | Code (binary) | Value (decimal) |
|-----------------|--------|---------------|-----------------|
| A               | A      | $0000_{(2)}$  | $0_{(10)}$      |
| C               | C      | $0001_{(2)}$  | $1_{(10)}$      |
| G               | G      | $0010_{(2)}$  | $2_{(10)}$      |
| T               | T      | $0011_{(2)}$  | $3_{(10)}$      |
| [AC]            | M      | $0100_{(2)}$  | $4_{(10)}$      |
| [AG]            | R      | $0101_{(2)}$  | $5_{(10)}$      |
| [AT]            | W      | $0110_{(2)}$  | $6_{(10)}$      |
| [CG]            | S      | $0111_{(2)}$  | $7_{(10)}$      |
| [CT]            | Y      | $1000_{(2)}$  | $8_{(10)}$      |
| [GT]            | K      | $1001_{(2)}$  | $9_{(10)}$      |
| [ACG]           | V      | $1010_{(2)}$  | $10_{(10)}$     |
| [ACT]           | H      | $1011_{(2)}$  | $11_{(10)}$     |
| [AGT]           | D      | $1100_{(2)}$  | $12_{(10)}$     |
| [CGT]           | B      | $1101_{(2)}$  | $13_{(10)}$     |
| [ACGT]          | N      | $1110_{(2)}$  | $14_{(10)}$     |
| -               | -      | $1111_{(2)}$  | $15_{(10)}$     |

**Table 3. List of all pair values.**

| p\t | A | C | G | T |
|---|---|---|---|---|
| A | 0 (0) | 1 (1) | 1 (2) | 1 (3) |
| C | 1 (4) | 0 (5) | 1 (6) | 1 (7) |
| G | 1 (8) | 1 (9) | 0 (10) | 1 (11) |
| T | 1 (12) | 1 (13) | 1 (14) | 0 (15) |
| M | 0 (16) | 0 (17) | 1 (18) | 1 (19) |
| R | 0 (20) | 1 (21) | 0 (22) | 1 (23) |
| W | 0 (24) | 1 (25) | 1 (26) | 0 (27) |
| S | 1 (28) | 0 (29) | 0 (30) | 1 (31) |
| Y | 1 (32) | 0 (33) | 1 (34) | 0 (35) |
| K | 1 (36) | 1 (37) | 0 (38) | 0 (39) |
| V | 0 (40) | 0 (41) | 0 (42) | 1 (43) |
| H | 0 (44) | 0 (45) | 1 (46) | 0 (47) |
| D | 0 (48) | 1 (49) | 0 (50) | 0 (51) |
| B | 1 (52) | 0 (53) | 0 (54) | 0 (55) |
| N | 0 (56) | 0 (57) | 0 (58) | 0 (59) |
| - | 1 (60) | 1 (61) | 1 (62) | 1 (63) |

**Table 4. Illustration of the proposed method.**

| $t_{(2)}$ | 00 | 11 | 10 | 00 | 01 | 01 | 10 | 10 | 01 | 00 | 11 |
|---|---|---|---|---|---|---|---|---|---|---|---|
| $i = 1, r_{(2)}$ | 000100 | 110100 | 001000 | 001000 | 011100 | | | | | | |
| $l_{(2)} = t_{(2)}$ OR $r_{(2)}$ | 000100 | 110111 | 001010 | 001000 | 011101 | | | | | | |
| $l_{(10)}$ | 4 | 55 | 10 | 8 | 29 | | | | | | |
| match | 1 | 0 | 0 | 1 | 0 | | | | | | |

| $t_{(2)}$ | 00 | 11 | 10 | 00 | 01 | 01 | 10 | 10 | 01 | 00 | 11 |
|---|---|---|---|---|---|---|---|---|---|---|---|
| $i = 2, r_{(2)}$ | | 000100 | 110100 | 001000 | 001000 | 011100 | | | | | |
| $l_{(2)} = t_{(2)}$ OR $r_{(2)}$ | | 000111 | 110110 | 001000 | 001001 | - | | | | | |
| $l_{(10)}$ | | 7 | 54 | 8 | 9 | - | | | | | |
| match | | 1 | 0 | 1 | 1 | - | | | | | |

| $t_{(2)}$ | 00 | 11 | 10 | 00 | 01 | 01 | 10 | 10 | 01 | 00 | 11 |
|---|---|---|---|---|---|---|---|---|---|---|---|
| $i = 3, r_{(2)}$ | | | 000100 | 110100 | 001000 | 001000 | 011100 | | | | |
| $l_{(2)} = t_{(2)}$ OR $r_{(2)}$ | | | 000110 | 110100 | 001001 | - | - | | | | |
| $(t_{(2)}$ OR $r_{(2)})_{(10)}$ | | | 6 | 52 | 9 | - | - | | | | |
| match | | | 1 | 1 | 1 | - | - | | | | |

| $t_{(2)}$ | 00 | 11 | 10 | 00 | 01 | 01 | 10 | 10 | 01 | 00 | 11 |
|---|---|---|---|---|---|---|---|---|---|---|---|
| $i = 4$, $r_{(2)}$ | | | | 000100 | 110100 | 001000 | 001000 | 011100 | | | |
| $l_{(2)} = t_{(2)}$ OR $r_{(2)}$ | | | | 000100 | 110101 | 001001 | 001010 | 011110 | | | |
| $l_{(10)}$ | | | | 4 | 53 | 9 | 10 | 30 | | | |
| match | | | | 1 | 0 | 1 | 0 | 0 | | | |

| $t_{(2)}$ | 00 | 11 | 10 | 00 | 01 | 01 | 10 | 10 | 01 | 00 | 11 |
|---|---|---|---|---|---|---|---|---|---|---|---|
| $i = 5$, $r_{(2)}$ | | | | 000100 | 110100 | 001000 | 001000 | 011100 | | | |
| $l_{(2)} = t_{(2)}$ OR $r_{(2)}$ | | | | 000101 | 110101 | 001010 | 001010 | 011101 | | | |
| $l_{(10)}$ | | | | 5 | 53 | 10 | 10 | 29 | | | |
| match | | | | 0 | 0 | 0 | 0 | 0 | | | |

| $t_{(2)}$ | 00 | 11 | 10 | 00 | 01 | 01 | 10 | 10 | 01 | 00 | 11 |
|---|---|---|---|---|---|---|---|---|---|---|---|
| $i = 6$, $r_{(2)}$ | | | | | 000100 | 110100 | 001000 | 001000 | 011100 | | |
| $l_{(2)} = t_{(2)}$ OR $r_{(2)}$ | | | | | 000101 | 110110 | 001010 | 001001 | 011100 | | |
| $l_{(10)}$ | | | | | 5 | 54 | 10 | 9 | 28 | | |
| match | | | | | 0 | 0 | 0 | 1 | 1 | | |

| $t_{(2)}$ | 00 | 11 | 10 | 00 | 01 | 01 | 10 | 10 | 01 | 00 | 11 |
|---|---|---|---|---|---|---|---|---|---|---|---|
| $i = 7$, $r_{(2)}$ | | | | | | | 000100 | 110100 | 001000 | 001000 | 011100 |
| $l_{(2)} = t_{(2)}$ OR $r_{(2)}$ | | | | | | | 000110 | 110110 | 001001 | 001000 | - |
| $l_{(10)}$ | | | | | | | 6 | 54 | 9 | 8 | - |
| match | | | | | | | 1 | 0 | 1 | 1 | - |

**Table 5.** Results for sequence $m = 500$.

| No | Position in sequence | Runtime [s] |
|---|---|---|
| 1. | 150967902 – 150967401 | 6.264 ± 0.056 |
| 2. | 52803498 – 52803997 | 6.784 ± 0.126 |
| 3. | 8749475 – 8749974 | 6.852 ± 0.072 |
| 4. | 234748299 – 234748798 | 6.870 ± 0.05 |
| 5. | 90472976 – 90473475 | 6.204 ± 0.086 |

**Table 6.** Results for sequence $m = 10000$.

| No | Position in sequence | Runtime [s] |
|---|---|---|
| 1. | 234097372 – 234107371 | 6.962 ± 0.038 |
| 2. | 33957613 – 33967612 | 7.054 ± 0.024 |
| 3. | 7034775 – 7044774 | 7.110 ± 0.03 |
| 4. | 115305600 – 115315599 | 6.272 ± 0.018 |
| 5. | 199419896 – 199429895 | 7.012 ± 0.012 |

**Table 7.** Results for sequence *m* = 100000.

| No | Position in sequence | Runtime [s] |
|---|---|---|
| 1. | 234097372 – 234197371 | 7.094 ± 0.026 |
| 2. | 33957613 – 34057612 | 6.374 ± 0.024 |
| 3. | 7034775 – 7135774 | 7.148 ± 0.028 |
| 4. | 115305600 – 115405599 | 7.190 ± 0.030 |
| 5. | 199419896 – 199519895 | 7.094 ± 0.026 |